\documentclass[10pt,twocolumn,letterpaper]{article}
\pdfoutput=1
\usepackage[pagenumbers]{cvpr} % To force page numbers, e.g. for an arXiv version

\usepackage{graphicx}
\usepackage{amsmath}
\usepackage{amssymb}
\usepackage{booktabs}
\usepackage{array}
\usepackage{color}
\usepackage{colortbl}
\usepackage{hhline}
\usepackage{soul}

\newcommand{\keypoint}[1]{\vspace{0.1cm}\noindent\textbf{#1}\quad}
\usepackage[pagebackref,breaklinks,colorlinks]{hyperref}
\usepackage{multirow}

\def\cvprPaperID{2} % *** Enter the CVPR Paper ID here
\def\confName{CVPR}
\def\confYear{2023}

\begin{document}

\title{An Evaluation of Non-Contrastive Self-Supervised Learning for Federated Medical Image Analysis}

\author{Soumitri Chattopadhyay\textsuperscript{*1} \hspace{.05cm} Soham Ganguly\textsuperscript{*1} \hspace{.05cm} Sreejit Chaudhury\textsuperscript{*1} \hspace{.05cm} Sayan Nag\textsuperscript{*2} \hspace{.05cm} Samiran Chattopadhyay\textsuperscript{1} \\
\textsuperscript{1}Jadavpur University \qquad\textsuperscript{2}University of Toronto\\
{\tt\small \{soumitri.chattopadhyay, sohamgangulysmail, sreejitchau, nagsayan112358\}@gmail.com} \\
{\tt\small samiran.chattopadhyay@jadavpuruniversity.in}
}

\newcommand{\snag}{\color{magenta}} 

% \and
\maketitle

%%%%%%%%% ABSTRACT
\begin{abstract}
  Privacy and annotation bottlenecks are two major issues that profoundly affect the practicality of machine learning-based medical image analysis. Although significant progress has been made in these areas, these issues are not yet fully resolved. In this paper, we seek to tackle these concerns head-on and systematically explore the applicability of non-contrastive self-supervised learning (SSL) algorithms under federated learning (FL) simulations for medical image analysis. We conduct thorough experimentation of recently proposed state-of-the-art non-contrastive frameworks under standard FL setups. With the SoTA Contrastive Learning algorithm, SimCLR as our comparative baseline, we benchmark the performances of our 4 chosen non-contrastive algorithms under non-i.i.d. data conditions and with a varying number of clients. We present a holistic evaluation of these techniques on 6 standardized medical imaging datasets. We further analyse different trends inferred from the findings of our research, with the aim to find directions for further research based on ours. To the best of our knowledge, ours is the first to perform such a thorough analysis of federated self-supervised learning for medical imaging. All of our source code will be made public upon acceptance of the paper.
\end{abstract}

\begin{figure*}[t]
\label{fig:teaser}
\includegraphics[width=\textwidth]{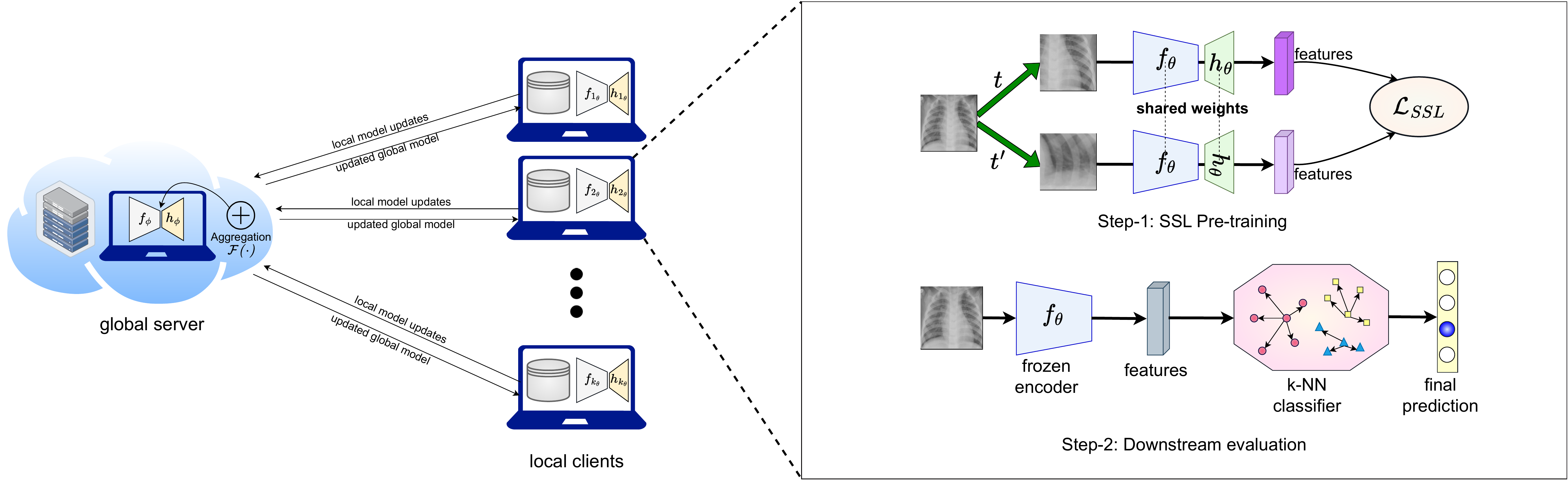}
\caption{Overall pipeline of the presented study. The SSL methods have been described in \autoref{fig:fig2ssl}.}
\label{fig:mainfig}
\end{figure*}

\def\thefootnote{*}\footnotetext{denotes equal contribution.}\def\thefootnote{\arabic{footnote}}

\section{Introduction}
% 1: FL and its importance
% 2. Medical imaging -- 2 aspects -- privacy/distributed and labeling
% 3. privacy/distributed handled by FL
% 4. labeling -- SSL -- how it is good at unsup learning
% 5. in this work -- bencharmking on medmnist -- FL+SSL. first of its kind

Medical image analysis \cite{aljuaid2022survey, litjens2017survey} is a topic of active research in the machine learning community. It involves an array of different tasks like disease detection \cite{manna2021interpretive}, classification \cite{azizi2021big, yousefi2019transfer} and segmentation \cite{basak2022exceedingly, basak2022addressing}. Classically, standard supervised computer vision techniques have been applied to solve medical imaging problems. Although such methods have achieved commendable performance across medical vision tasks, there are two unique challenges the field faces, which are yet to be fully tackled by such methods. 

The first major challenge in practical medical imaging is learning over medical data stored in a distributed manner. In the field of medical imaging, the entire data is rarely stored in a centralized server; it is distributed across various servers and devices (say, servers of different hospitals containing patient data). Data stored at different sites may even have very different distributions \cite{degan2022application}. Furthermore, owing to privacy and legal concerns, such data can neither be accumulated together into a centralised server nor be shared with other sources for training deep learning models. Thus, one requires decentralized training across all such client devices that preserves data privacy. This has been facilitated by the paradigm of federated learning (FL) -- introduced in the seminal work by \cite{mcmahan2017communication}. FL involves individual model training in each client, followed by aggregating the individual model weights into a global server model, whose copies are then sent to the clients for inference. This is non-trivial, since the server model would have to fit different data distributions across the clients. 

Secondly, annotating medical imaging datasets is an arduous task, requiring a significant amount of time and effort from skilled clinicians. In fact, this is a major reason why only a handful of large-scale labeled medical imaging datasets are available to the community. This has led to the development of annotation-efficient paradigms such as semi-supervised \cite{yang2021survey, basak2022ideal}, weakly-supervised \cite{rony2019deep} and most notably, self-supervised learning (SSL) algorithms. In particular, SSL \cite{jing2020self, chen2020simple, bardes2021vicreg, manna2022swis, zhu2022tico} has shown great potential to learn robust representations in computer vision \cite{jaiswal2020survey, jing2020self, manna2022swis} and medical imaging \cite{azizi2021big, manna2021interpretive, shurrab2022self} tasks, achieving comparable performance to even supervised approaches.

In this study, we aim to tackle the two aforementioned problems in the field of medical imaging head-on by leveraging SSL to federated setups for medical imaging. In particular, we conduct a thorough investigation of the performances of SoTA \emph{non-contrastive} SSL algorithms under various federated learning setups. The motivation for our focus on non-contrastive methods \cite{chen2021exploring, bardes2021vicreg, zbontar2021barlow, zhu2022tico} is in light of their using lower batch sizes for reducing computational demands. To evaluate their performance, we utilise the SoTA contrastive learning algorithm SimCLR \cite{chen2020simple} as a comparative baseline. On the federated side, we conduct experiments using three standard FL algorithms, namely FedAVG \cite{mcmahan2017communication}, FedBN \cite{li2021fedbn} and FedProx \cite{li2020fedprox}. By varying the number of clients, we put forth a rigorous evaluation of such simulations on datasets of the standardized medical imaging suite, MedMNIST \cite{yang2021medmnist, yang2021medmnistv2}, which can be used as a ready reference for current and future research, something that has not been done so far. Moreover, we further discuss the various trends observed in the evaluation, which could provide intriguing insights and pave the way for further study. To the best of our knowledge, such a holistic analysis has not been performed in literature to date, making our contribution significant. We believe such a study would be very helpful to the medical vision community in a very practical manner.

\begin{figure*}[t]
    \centering
    \includegraphics[width=0.95\linewidth]{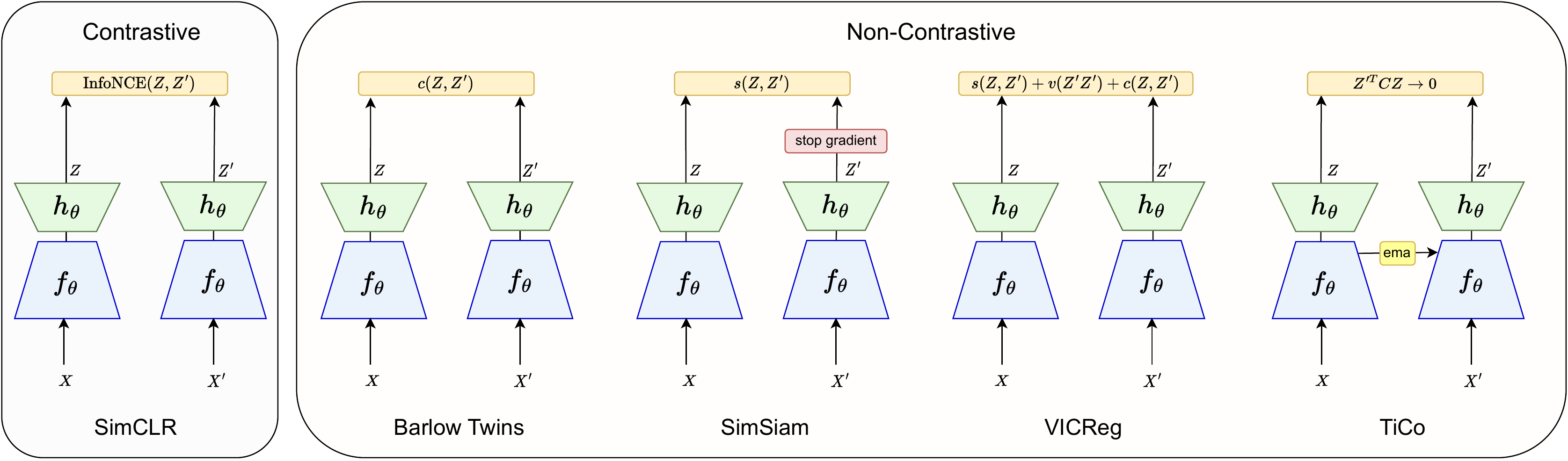}
    \caption{Comparison of different SSL methods used in the presented study.}
    \label{fig:fig2ssl}
\end{figure*}

% Our codes will be made public

To sum up, the contributions of our work are as follows:

\begin{enumerate}
    \item We present the first thorough investigation of the applicability of non-contrastive self-supervision algorithms under three different federated learning setups in literature for medical image analysis.
    \item We perform a rigorous evaluation of the performance of these algorithms under non-i.i.d. data splits with variation in the number of clients participating in federated learning.
    \item The benchmarking is performed on the MedMNIST suite, which consists of standardized medical image datasets. Furthermore, we systematically analyse different trends inferred from the study, paving the way for further research.
\end{enumerate}

Our experimental pipeline is schematically represented in \autoref{fig:mainfig}. All source codes will be made publicly available upon acceptance.

\section{Related Work}

\keypoint{Federated Learning:}FL \cite{mcmahan2017communication, li2020fedprox, zhao2018FedAVGshare, li2021fedbn, liu2021feddg} aims at training machine learning models across a number of private devices having their own datasets, with the underlying assumption that the clients' data are mutually independent of each other \cite{mcmahan2017communication}. In other words, FL involves decentralized training of models in a distributed manner. For generalization, the individually trained models are aggregated at a global server, via an aggregation function. Thus, no data is shared among the clients, only the model parameters being communicated in the aggregation process \cite{qu2022rethinking}. First proposed in \cite{mcmahan2017communication}, the classic FedAVG algorithm, which performs a simple weighted averaging of model weights across clients, is still considered a strong baseline framework. Among other popular FL methods, FedProx \cite{li2020fedprox} introduced a proximal term to the local client losses so as to mitigate possible weight divergences, FedAVG-Share \cite{zhao2018FedAVGshare} proposed to use a small globally shared data subset for robust generalization, and FedBN \cite{li2021fedbn} used local batch normalizations to capture different data distributions and alleviate representational shift prior to aggregation. In this paper, we adopted FedAVG \cite{mcmahan2017communication}, FedBN \cite{li2021fedbn} and FedProx \cite{li2020fedprox} as for conducting the FL simulations over SSL algorithms.

\keypoint{Self-Supervised Learning:}Being a subset of unsupervised learning, SSL aims at learning representations without the need for explicit annotations by means of pre-training tasks in which the so-called pseudo-ground-truths are generated from the data itself. While classical approaches include colorization \cite{zhang2016colorful}, inpainting \cite{pathak2016context} and solving jigsaw puzzles \cite{noroozi2016unsupervised}, recent SoTA approaches have focused on contrastive learning \cite{chen2021intriguing, chen2020simple, he2020momentum}, which aims at pulling similar data points (positive samples) closer together, along with pushing apart dissimilar points (negative samples) in the embedding space. Although popular contrastive learning frameworks such as SimCLR \cite{chen2020simple} and MoCo \cite{he2020momentum} have proven to be successful at learning robust visual representations, a bottleneck they face is the requirement of a large number of negative samples in a mini-batch, which compels the use of large batch sizes. To alleviate this limitation, more recent SSL works have directed towards the paradigm of non-contrastive learning \cite{zbontar2021barlow, zhu2022tico, bardes2021vicreg} -- wherein the objective is to solely bring similar entities closer together in the latent space, thereby mitigating the need for negative samples altogether. Barlow Twins \cite{zbontar2021barlow} uses information maximization to maximize similarity along with reducing redundancy among neurons, inspired by neuroscience theories. VICReg \cite{bardes2021vicreg} extended upon the former by introducing a hinge function for variance regularization. In our study, we adopted some of the recently proposed non-contrastive SSL variants -- SimSiam \cite{chen2021exploring}, Barlow Twins \cite{zbontar2021barlow} and VICReg \cite{bardes2021vicreg}. We also considered TiCo \cite{zhu2022tico}, which combines the flavours of both contrastive and non-contrastive learning along with the use of an implicit feature memory. For evaluation purpose, SimCLR \cite{chen2020simple} was chosen as the contrastive baseline method. 

\section{Federated Self-Supervised Learning}

As explained, our study investigates self-supervision models under FL simulations. In this section, we provide background on the SSL and federated learning algorithms used in this paper.

\subsection{SSL Methods}

\keypoint{Contrastive:}We treat contrastive SSL as a baseline and employ the SoTA SimCLR \cite{chen2020simple} algorithm, which uses the InfoNCE \cite{chen2021intriguing, oord2018representation, kukleva2023temperature} contrastive loss to push embeddings of different views of the same image closer and pull apart those of different images.

\keypoint{Non-contrastive:}As our primary focus is towards non-contrastive SSL, we experiment with several such methods, namely SimSiam \cite{chen2021exploring}, Barlow Twins \cite{zbontar2021barlow}, VICReg \cite{bardes2021vicreg} and TiCo \cite{zhu2022tico}. It is worth noting that TiCo can be interpreted as both contrastive and non-contrastive; it has  elements of both the SSL variants. 

For a high-level understanding, the SSL methods adopted in this study have been depicted figuratively in \autoref{fig:fig2ssl}. Other details have been provided in \autoref{tab:hp}. 

\subsection{FL Algorithms}
We simulate three standard FL algorithms in our study, namely FedAVG \cite{mcmahan2017communication}, FedBN \cite{li2021fedbn} and FedProx \cite{li2020fedprox}.

\keypoint{FedAVG:}FedAVG utilizes iterative averaging over client weights to update the global model \cite{mcmahan2017communication}. In each communication round, a set of selected clients $S_{t}$ receive the current global model, locally train for a predetermined \textit{E} epochs, and send the local weights back to the central server. The server updates the global model by averaging the client weights weighted by number of local samples $n_{k}$ at each client.
\begin{gather}
w_{t+1} \leftarrow \sum_{k\mathcal{E}S_{t}}\frac{n_{k}}{n}w^{k}_{t+1} 
\end{gather}
Here, $w_{t+1}$ represents the updated global model in round $t+1$, $w^{k}_{t+1}$ are the weights received from the $k^{th}$ client and $n$ is the total number of samples present in selected clients.

\keypoint{FedBN:}FedBN is a modification of the FedAVG algorithm designed to address the different marginal or conditional feature distributions across clients, a problem known as feature shift non-i.i.d. \cite{li2021fedbn}. FedBN utilizes local batch normalization, with updates to global model weights via weighted averaging, \emph{leaving out} the weights from the batch normalization layers.

\keypoint{FedProx:}Proposed in \cite{li2020fedprox}, FedProx is a generalisation of FedAVG with modifications intended to address heterogeneous data and systems. Like FedAVG, in each communication round, the server selects a set of clients and sends them the current global model $w^{t}$. Unlike FedAVG, FedProx allows local optimization on clients to run for a variable number of epochs by making clients optimise a regularised loss with a proximal term. Each client approximately minimizes the a new object $h_{k}$ defined over its original local objective $F_{k}$ as (Here, $\mu$ is the regularization hyperparameter):
\begin{gather}
\min_{w}h_{k}(w;w^{t}) = F_{k} + \frac{\mu}{2}\|w-w^{t}\|^{2}
\end{gather}

\section{Experimental Protocol}

\subsection{Datasets}\label{sec:data}
We evaluate the federated self-supervised simulations on the MedMNIST data suite \cite{yang2021medmnist, yang2021medmnistv2}, a large-scale collection of standardized real-world biomedical images resized to MNIST-like dimensions. The datasets used in this work are \textbf{Pneumonia, Breast, Retina, Organ-A, Organ-C and Organ-S}. It is essential to mention that \{RetinaMNIST, BreastMNIST and PneumoniaMNIST\} are ``\textit{small datasets}'' (having less than 6k images in total), while the remaining are different views (A=axial, C=coronal, S=sagittal) of abdominal CT images, each having more than 20k images, thereby being ``\textit{big datasets}''. More details about the datasets can be found in \footnote{\href{https://medmnist.com/}{https://medmnist.com/}}.

\subsection{Configuration}\label{sec:config}
We conducted our experiments in PyTorch \cite{paszke2019pytorch} accelerated by a 16GB NVIDIA Tesla V100 GPU. For self-supervised model training, a ResNet-18 \cite{he2016residual} encoder was used with random initialization, which is considered a de facto standard for SSL works \cite{chen2020simple, zbontar2021barlow}. For the projector head, a $128$-dimensional fully connected (FC) output layer was considered for SimCLR, while a $512$-dimensional FC layer was used as the expander for the non-contrastive learning algorithms. Following standard works, we set the number of clients to $5, 10$ and $20$ for our investigations. For creating class imbalance in our clients to simulate the non-i.i.d. condition of FL, we utilised Dirichlet distribution \cite{mcmahan2017communication, li2021fedbn} to generate data splits. We ran each experiment for a total of 20 communication rounds with each client iterating over 20 internal epochs in each round. All hyperparameters for our experiments have been described in \autoref{tab:hp}, and were fixed across experiments for a fair evaluation. It should be mentioned that we could not go beyond a batch size of $128$ due to computational constraints. 

\begin{table}[!h]
\centering
\resizebox{\columnwidth}{!}{
\begin{tabular}{c|c} 
\hline
\multicolumn{2}{c}{\textbf{SSL Pre-training}}  
\\ \hline
backbone                  & ResNet18    \\ 
learning rate             & $0.01$      \\ 
batch size ($\#clients\in\{5,10\}$)               & $128$       \\
batch size ($\#clients=20$)               & $64$ \\  
epochs per FL round           & $20$        \\ 
temperature (SimCLR)      & $\tau=0.5$     
\\ \hline
\multicolumn{2}{c}{\textbf{FL setting}}  
\\ \hline
$\#clients$                 & $5, 10, 20$ \\ 
rounds                    & $20$        \\ 
non-i.i.d. split             & Dirichlet ($\alpha=0.1$)  \\ 
Proximal weight (FedProx) & $\mu=0.001$                
\\ \midrule
\multicolumn{2}{c}{\textbf{KNN evaluation}}  
\\ \hline
distance metric                    & Euclidean distance          \\ 
$k$                     & $20$      
\\ \hline
\end{tabular}
}
\caption{Hyperparameter setting or our experimental protocol.}
\label{tab:hp}
\end{table}

For evaluation purposes, following recent SSL literature \cite{yeh2022decoupled, kukleva2023temperature} we used a simple KNN classification protocol ($k=20$) with Euclidean distance metric \cite{laaksonen1996classification} on the representations obtained by the frozen encoder network. This offers a simple benchmarking scheme without many bells and whistles directly on features learnt in a completely unsupervised manner.

% \keypoint{\blue{Hyperparameters:}} \blue{ We utilized a learning rate of 0.01 in each of our clients and fixed batch size to 128 samples for every algorithm. For the FedProx algorithm, we set the parameter $\mu=0.001$.The Dirichlet distribution used to sample the data had parameter $\alpha=0.1$. For KNN evaluation, experimentation was performed with k set to values of 10, 20, 50 and 100 with the best performance being obtained at k=20. Accordingly, the paper contains results obtained for each algorithm with k set to 20.
% }

\subsection{Evaluation metrics}
Two standard classification metrics have been used for the performance evaluation of the algorithms investigated by this study: \textit{Accuracy} and weighted \textit{F1-Score}. For reporting empirical results, we used F1-score and for qualitative analyses, we used accuracy values. We report the average over individual client scores for all experiments.

\begin{table*}[h]
\resizebox{\textwidth}{!}{
    \begin{tabular}{c|ccc|ccc|ccc}
% \multicolumn{10}{c}{Pneumonia}           
\toprule
\multirow{2}{*}{\textbf{SSL Method}} & \multicolumn{3}{c|}{$\#clients=5$}       & \multicolumn{3}{c|}{$\#clients=10$}      & \multicolumn{3}{c}{\textbf{$\#clients=20$}}      \\
                               & \textbf{FedAVG}      & \textbf{FedBN}       & \textbf{FedProx}     & \textbf{FedAVG}      & \textbf{FedBN}       & \textbf{FedProx}     & \textbf{FedAVG}      & \textbf{FedBN}       & \textbf{FedProx}    \\ \midrule
\textbf{Barlow}                  & 0.837{ $\pm$ }0.148 & 0.746{ $\pm$ }0.240 & 0.854{ $\pm$ }0.133 & 0.831{ $\pm$ }0.149    & 0.753{ $\pm$ }0.221 & 0.810{ $\pm$ }0.174 & \textbf{0.826{ $\pm$ }0.147}    & 0.770{ $\pm$ }0.199 & \textbf{0.829{ $\pm$ }0.144}  \\
\textbf{VICReg}                  & \textbf{0.869{ $\pm$ }0.121}   & 0.844{ $\pm$ }0.144 & 0.878{ $\pm$ }0.113   & \textbf{0.872{ $\pm$ }0.117}    & \textbf{0.837{ $\pm$ }0.145} & 0.870{ $\pm$ }0.116 & 0.819{ $\pm$ }0.145    & \textbf{0.810{ $\pm$ }0.162} & 0.826{ $\pm$ }0.145 \\
\textbf{SimSiam}                 & 0.771{ $\pm$ }0.211   & 0.776{ $\pm$ }0.206   & 0.771{ $\pm$ }0.211 & 0.806{ $\pm$ }0.170    & 0.774{ $\pm$ }0.200   & 0.807{ $\pm$ }0.168  & 0.773{ $\pm$ }0.199    & 0.779{ $\pm$ }0.188  & 0.774{ $\pm$ }0.192 \\
\textbf{TiCo}                    & 0.826{ $\pm$ }0.162   & 0.798{ $\pm$ }0.186   & 0.825{ $\pm$ }0.164 & 0.792{ $\pm$ }0.183    & 0.779{ $\pm$ }0.196   & 0.826{ $\pm$ }0.154 & 0.772{ $\pm$ }0.195    & 0.768{ $\pm$ }0.196 & 0.770{ $\pm$ }0.197   \\ \midrule
\textbf{SimCLR}                  & 0.867{ $\pm$ }0.124   & \textbf{0.851{ $\pm$ }0.140}   & \textbf{0.879{ $\pm$ }0.112} & 0.871{ $\pm$ }0.113    & 0.829{ $\pm$ }0.151   & \textbf{0.878{ $\pm$ }0.108} & 0.819{ $\pm$ }0.153    & 0.781{ $\pm$ }0.187 & 0.819{ $\pm$ }0.155   \\ \bottomrule
\end{tabular}
}
\caption{$F1-scores$ obtained on Pneumonia-MNIST dataset.}
\label{tab:pneumonia}
\end{table*}

\begin{table*}[h]
\resizebox{\textwidth}{!}{
    \begin{tabular}{c|ccc|ccc|ccc}
% \multicolumn{10}{c}{Breast}           
\toprule
\multirow{2}{*}{\textbf{SSL Method}} & \multicolumn{3}{c|}{$\#clients=5$}       & \multicolumn{3}{c|}{$\#clients=10$}      & \multicolumn{3}{c}{\textbf{$\#clients=20$}}      \\
                               & \textbf{FedAVG}      & \textbf{FedBN}       & \textbf{FedProx}     & \textbf{FedAVG}      & \textbf{FedBN}       & \textbf{FedProx}     & \textbf{FedAVG}      & \textbf{FedBN}       & \textbf{FedProx}    \\ \midrule
\textbf{Barlow}               & 0.755{ $\pm$ }0.225   & \textbf{0.785{ $\pm$ }0.197} & 0.755{ $\pm$ }0.225   & \textbf{0.823{ $\pm$ }0.181}    & 0.769{ $\pm$ }0.258 & \textbf{0.822{ $\pm$ }0.180} & 0.675{ $\pm$ }0.319    & \textbf{0.724{ $\pm$ }0.291} & 0.675{ $\pm$ }0.319 \\
\textbf{VICReg}               & \textbf{0.757{ $\pm$ }0.228}   & 0.764{ $\pm$ }0.218 & \textbf{0.757{ $\pm$ }0.228}   & 0.803{ $\pm$ }0.197    & 0.783{ $\pm$ }0.223 & 0.802{ $\pm$ }0.197 & \textbf{0.711{ $\pm$ }0.284}    & 0.678{ $\pm$ }0.31  & \textbf{0.711{ $\pm$ }0.284} \\
\textbf{SimSiam}              & 0.731{ $\pm$ }0.250   & 0.737{ $\pm$ }0.256 & 0.731{ $\pm$ }0.250   & 0.722{ $\pm$ }0.287    & 0.767{ $\pm$ }0.247 & 0.721{ $\pm$ }0.287 & 0.702{ $\pm$ }0.294    & 0.680{ $\pm$ }0.307 & 0.702{ $\pm$ }0.294 \\
\textbf{TiCo}                 & 0.731{ $\pm$ }0.215   & 0.766{ $\pm$ }0.215 & 0.731{ $\pm$ }0.248   & 0.748{ $\pm$ }0.267    & 0.770{ $\pm$ }0.240 & 0.747{ $\pm$ }0.266 & 0.696{ $\pm$ }0.295    & 0.676{ $\pm$ }0.312 & 0.696{ $\pm$ }0.294 \\ \midrule
\textbf{SimCLR}               & 0.727{ $\pm$ }0.264   & 0.776{ $\pm$ }0.215 & 0.727{ $\pm$ }0.264 & 0.793{ $\pm$ }0.223    & \textbf{0.786{ $\pm$ }0.232} & 0.792{ $\pm$ }0.222 & 0.695{ $\pm$ }0.295    & 0.693{ $\pm$ }0.292 & 0.695{ $\pm$ }0.295 \\ \bottomrule
\end{tabular}
}
\caption{$F1-scores$ obtained on Breast-MNIST dataset.}
\label{tab:breast}
\end{table*}

\begin{table*}[!h]
\resizebox{\textwidth}{!}{
    \begin{tabular}{c|ccc|ccc|ccc}
% \multicolumn{10}{c}{Retina}           
\toprule
\multirow{2}{*}{\textbf{SSL Method}} & \multicolumn{3}{c|}{$\#clients=5$}       & \multicolumn{3}{c|}{$\#clients=10$}      & \multicolumn{3}{c}{\textbf{$\#clients=20$}}      \\
                               & \textbf{FedAVG}      & \textbf{FedBN}       & \textbf{FedProx}     & \textbf{FedAVG}      & \textbf{FedBN}       & \textbf{FedProx}     & \textbf{FedAVG}      & \textbf{FedBN}       & \textbf{FedProx}    \\ \midrule
\textbf{Barlow}               & 0.497{ $\pm$ }0.302   & 0.494{ $\pm$ }0.300 & 0.500{ $\pm$ }0.294 & 0.486{ $\pm$ }0.280    & 0.494{ $\pm$ }0.261 & 0.486{ $\pm$ }0.280 & 0.470{ $\pm$ }0.291    & 0.465{ $\pm$ }0.280 & 0.470{ $\pm$ }0.291 \\
\textbf{VICReg}               & \textbf{0.535{ $\pm$ }0.265}   & 0.505{ $\pm$ }0.300 & 0.518{ $\pm$ }0.279 & \textbf{0.506{ $\pm$ }0.250}    & 0.497{ $\pm$ }0.26  & \textbf{0.505{ $\pm$ }0.249} & \textbf{0.474{ $\pm$ }0.269}    & 0.478{ $\pm$ }0.272 & \textbf{0.474{ $\pm$ }0.269} \\
\textbf{SimSiam}              & 0.499{ $\pm$ }0.299   & 0.504{ $\pm$ }0.294 & 0.523{ $\pm$ }0.277 & 0.486{ $\pm$ }0.270    & 0.498{ $\pm$ }0.258 & 0.485{ $\pm$ }0.27  & 0.451{ $\pm$ }0.268    & 0.468{ $\pm$ }0.255 & 0.451{ $\pm$ }0.268 \\
\textbf{TiCo}                 & 0.519{ $\pm$ }0.280   & 0.503{ $\pm$ }0.297 & \textbf{0.540{ $\pm$ }0.261} & 0.500{ $\pm$ }0.258    & 0.497{ $\pm$ }0.256 & 0.500{ $\pm$ }0.257 & 0.448{ $\pm$ }0.271    & \textbf{0.479{ $\pm$ }0.266} & 0.447{ $\pm$ }0.270 \\ \midrule
\textbf{SimCLR}               & 0.522{ $\pm$ }0.277   & \textbf{0.513{ $\pm$ }0.286} & 0.518{ $\pm$ }0.280 & 0.478{ $\pm$ }0.281    & \textbf{0.505{ $\pm$ }0.250} & 0.477{ $\pm$ }0.280 & 0.458{ $\pm$ }0.266    & 0.478{ $\pm$ }0.253 & 0.458{ $\pm$ }0.266 \\ \bottomrule
\end{tabular}
}
\caption{$F1-scores$ obtained on Retina-MNIST dataset.}
\label{tab:retina}
\end{table*}

\section{Results and Discussion}

We report the findings of the experimentation across the datasets as metrics averaged across clients, along with their standard deviation values. As discussed earlier, our experiments comprise evaluating different SSL algorithms over three popular FL setups, simulating variation in the number of clients. The exhaustive results for each dataset have been tabulated in tables \ref{tab:pneumonia}, \ref{tab:breast}, \ref{tab:retina}, \ref{tab:organA}, \ref{tab:organC} and \ref{tab:organS}.

\subsection{Across varying dataset sizes}
The six datasets utilised in our work fall into two categories: the $3$ \textit{small datasets}, Pneumonia ($\approx5.8$k images), Breast (780 images) and Retina (1.6k images); and the $3$ \textit{big datasets}, Organ-A ($\approx58$k images), Organ-C ($\approx23$k images), and Organ-S ($\approx25$k images). To investigate further into the effect of dataset size  on SSL algorithms under a federated setup, we compare the performance of the different SSL methods under study over these two categories of datasets. We analyze the algorithms both in terms of performance (mean F1-score) as well as stability (standard deviation of scores across clients).

% In the 3 big datasets, a clear trend emerges of algorithm performance deteriorating with increase in number of clients. In contrast, in the 3 small datasets, a majority of the algorithms peak in performance when utilising 10 clients. The performance then deteriorates when the number of clients is further increased further to 20, which can be attributed to the decrease in the local client dataset sizes hurting the learning ability of the SSL algorithm. Another notable observation in this regard is the performance of the FedBN algorithm with respect to FedAVG and FedProx. For small datasets the performances are in correspondence. However with the large sized datasets, FedBN tends to perform poorly in comparison.  

\begin{table*}[h]
\resizebox{\textwidth}{!}{
    \begin{tabular}{c|ccc|ccc|ccc}
% \multicolumn{10}{c}{OrganA}           
\toprule
\multirow{2}{*}{\textbf{SSL Method}} & \multicolumn{3}{c|}{$\#clients=5$}       & \multicolumn{3}{c|}{$\#clients=10$}      & \multicolumn{3}{c}{\textbf{$\#clients=20$}}      \\
                               & \textbf{FedAVG}      & \textbf{FedBN}       & \textbf{FedProx}     & \textbf{FedAVG}      & \textbf{FedBN}       & \textbf{FedProx}     & \textbf{FedAVG}      & \textbf{FedBN}       & \textbf{FedProx}    \\ \midrule
\textbf{Barlow}                         & 0.677{ $\pm$ }0.093 & 0.310{ $\pm$ }0.072 & 0.681{ $\pm$ }0.086 & 0.606{ $\pm$ }0.048 & 0.362{ $\pm$ }0.062 & 0.605{ $\pm$ }0.048 & 0.540{ $\pm$ }0.047 & 0.360{ $\pm$ }0.070 & 0.533{ $\pm$ }0.047 \\
\textbf{VICReg}                         & \textbf{0.721{ $\pm$ }0.085} & 0.414{ $\pm$ }0.088 & \textbf{0.724{ $\pm$ }0.089} & \textbf{0.681{ $\pm$ }0.052} & 0.399{ $\pm$ }0.048 & \textbf{0.682{ $\pm$ }0.055} & \textbf{0.630{ $\pm$ }0.044} & \textbf{0.512{ $\pm$ }0.054} & \textbf{0.623{ $\pm$ }0.046} \\
\textbf{SimSiam}                        & 0.441{ $\pm$ }0.026 & 0.455{ $\pm$ }0.087 & 0.441{ $\pm$ }0.081 & 0.406{ $\pm$ }0.059 & 0.426{ $\pm$ }0.052 & 0.404{ $\pm$ }0.056 & 0.405{ $\pm$ }0.061 & 0.428{ $\pm$ }0.059 & 0.402{ $\pm$ }0.062 \\
\textbf{TiCo}                           & 0.643{ $\pm$ }0.082 & \textbf{0.534{ $\pm$ }0.084} & 0.647{ $\pm$ }0.081 & 0.609{ $\pm$ }0.034 & 0.512{ $\pm$ }0.045 & 0.603{ $\pm$ }0.041 & 0.501{ $\pm$ }0.049 & 0.459{ $\pm$ }0.054 & 0.478{ $\pm$ }0.044 \\\midrule
\textbf{SimCLR}                         & 0.653{ $\pm$ }0.067 & 0.513{ $\pm$ }0.085 & 0.697{ $\pm$ }0.086 & 0.669{ $\pm$ }0.046 & \textbf{0.519{ $\pm$ }0.048} & 0.661{ $\pm$ }0.051 & 0.588{ $\pm$ }0.050 & 0.483{ $\pm$ }0.059 & 0.583{ $\pm$ }0.047    \\ \bottomrule
\end{tabular}
}
\caption{$F1-scores$ obtained on OrganA-MNIST dataset.}
\label{tab:organA}
\end{table*}

\begin{table*}[h]
\resizebox{\textwidth}{!}{
    \begin{tabular}{c|ccc|ccc|ccc}
% \multicolumn{10}{c}{OrganC}           
\toprule
\multirow{2}{*}{\textbf{SSL Method}} & \multicolumn{3}{c|}{$\#clients=5$}       & \multicolumn{3}{c|}{$\#clients=10$}      & \multicolumn{3}{c}{\textbf{$\#clients=20$}}      \\
                               & \textbf{FedAVG}      & \textbf{FedBN}       & \textbf{FedProx}     & \textbf{FedAVG}      & \textbf{FedBN}       & \textbf{FedProx}     & \textbf{FedAVG}      & \textbf{FedBN}       & \textbf{FedProx}    \\ \midrule
\textbf{Barlow}                & 0.648{ $\pm$ }0.069   & 0.291{ $\pm$ }0.061 & 0.644{ $\pm$ }0.081 & 0.593{ $\pm$ }0.076    & 0.327{ $\pm$ }0.059 & 0.575{ $\pm$ }0.078 & 0.559{ $\pm$ }0.067    & 0.320{ $\pm$ }0.084 & 0.558{ $\pm$ }0.066 \\
\textbf{VICReg}                & \textbf{0.696{ $\pm$ }0.072}   & 0.408{ $\pm$ }0.080 & \textbf{0.692{ $\pm$ }0.070} & \textbf{0.655{ $\pm$ }0.076}    & 0.429{ $\pm$ }0.073 & \textbf{0.653{ $\pm$ }0.080} & \textbf{0.594{ $\pm$ }0.071}    & \textbf{0.466{ $\pm$ }0.082} & \textbf{0.600{ $\pm$ }0.074} \\
\textbf{SimSiam}               & 0.376{ $\pm$ }0.077   & 0.374{ $\pm$ }0.075 & 0.376{ $\pm$ }0.078 & 0.384{ $\pm$ }0.079    & 0.395{ $\pm$ }0.072 & 0.373{ $\pm$ }0.090 & 0.367{ $\pm$ }0.089    & 0.389{ $\pm$ }0.082 & 0.367{ $\pm$ }0.090 \\
\textbf{TiCo}                  & 0.599{ $\pm$ }0.073   & 0.463{ $\pm$ }0.074 & 0.600{ $\pm$ }0.072 & 0.570{ $\pm$ }0.061    & 0.471{ $\pm$ }0.072 & 0.572{ $\pm$ }0.060 & 0.439{ $\pm$ }0.076    & 0.414{ $\pm$ }0.080 & 0.436{ $\pm$ }0.072 \\ \midrule
\textbf{SimCLR}                & 0.611{ $\pm$ }0.132   & \textbf{0.495{ $\pm$ }0.081} & 0.664{ $\pm$ }0.066 & 0.635{ $\pm$ }0.070    & \textbf{0.523{ $\pm$ }0.066} & 0.626{ $\pm$ }0.066 & 0.544{ $\pm$ }0.069    & 0.440{ $\pm$ }0.084 & 0.543{ $\pm$ }0.062 \\ \bottomrule
\end{tabular}
}
\caption{$F1-scores$ obtained on OrganC-MNIST dataset.}
\label{tab:organC}
\end{table*}

\begin{table*}[!h]
\resizebox{\textwidth}{!}{
    \begin{tabular}{c|ccc|ccc|ccc}
% \multicolumn{10}{c}{OrganS}           
\toprule
\multirow{2}{*}{\textbf{SSL Method}} & \multicolumn{3}{c|}{$\#clients=5$}       & \multicolumn{3}{c|}{$\#clients=10$}      & \multicolumn{3}{c}{\textbf{$\#clients=20$}}      \\
                               & \textbf{FedAVG}      & \textbf{FedBN}       & \textbf{FedProx}     & \textbf{FedAVG}      & \textbf{FedBN}       & \textbf{FedProx}     & \textbf{FedAVG}      & \textbf{FedBN}       & \textbf{FedProx}    \\ \midrule
\textbf{Barlow}                & 0.556{ $\pm$ }0.096   & 0.298{ $\pm$ }0.090 & 0.541{ $\pm$ }0.096  & 0.545{ $\pm$ }0.075    & 0.352{ $\pm$ }0.079 & 0.554{ $\pm$ }0.071 & 0.498{ $\pm$ }0.081    & 0.332{ $\pm$ }0.091 & 0.500{ $\pm$ }0.081 \\
\textbf{VICReg}                & \textbf{0.602{ $\pm$ }0.102}   & 0.365{ $\pm$ }0.089 & \textbf{0.607{ $\pm$ }0.094}  & 0.554{ $\pm$ }0.100    & 0.401{ $\pm$ }0.078 & \textbf{0.577{ $\pm$ }0.088} & \textbf{0.526{ $\pm$ }0.085}    & \textbf{0.423{ $\pm$ }0.073} & \textbf{0.531{ $\pm$ }0.082} \\
\textbf{SimSiam}               & 0.368{ $\pm$ }0.100   & 0.355{ $\pm$ }0.100 & 0.365{ $\pm$ }0.099  & 0.380{ $\pm$ }0.086    & 0.387{ $\pm$ }0.076 & 0.380{ $\pm$ }0.092 & 0.351{ $\pm$ }0.088    & 0.374{ $\pm$ }0.085 & 0.351{ $\pm$ }0.086 \\
\textbf{TiCo}                  & 0.519{ $\pm$ }0.097   & 0.415{ $\pm$ }0.087 & 0.510{ $\pm$ }0.102 & 0.502{ $\pm$ }0.062    & 0.431{ $\pm$ }0.073 & 0.502{ $\pm$ }0.068 & 0.395{ $\pm$ }0.085    & 0.387{ $\pm$ }0.080 & 0.413{ $\pm$ }0.091 \\ \midrule
\textbf{SimCLR}                & 0.527{ $\pm$ }0.104   & \textbf{0.428{ $\pm$ }0.105} & 0.560{ $\pm$ }0.099  & \textbf{0.578{ $\pm$ }0.065}    & \textbf{0.461{ $\pm$ }0.078} & 0.573{ $\pm$ }0.064 & 0.480{ $\pm$ }0.088    & 0.399{ $\pm$ }0.087 & 0.471{ $\pm$ }0.084 \\ \bottomrule
\end{tabular}
}
\caption{$F1-scores$ obtained on OrganS-MNIST dataset.}
\label{tab:organS}
\end{table*}

\keypoint{Small datasets:}Tables \ref{tab:pneumonia}, \ref{tab:breast} and \ref{tab:retina} contain the mean and standard deviations of weighted $F1-scores$ over individual clients for each SSL method over Pneumonia, Breast and Retina datasets respectively. From \autoref{tab:pneumonia} we can see that the best performer overall on Pneumonia is the non-contrastive method VICReg. While SimCLR gets highest scores in some cases under smaller number of clients, its performance significantly deteriorates as the number of clients is increased to $20$. In contrast, the performances of the non-contrastive methods, especially VICReg and Barlow Twins, decrease to a far smaller extent, making them the best performers under higher numbers of clients. When setting number of clients to $20$, the batch size is reduced to $64$ (to allow for lower per-client data availability due to increased number of clients). Contrastive learning methods like SimCLR are heavily dependent on the number of negative samples available, and thus perform better under higher batch sizes \cite{chen2020simple, chen2021intriguing}, while non-contrastive methods, not having such a dependency, are not affected as much. This trend is more visibly observable in \autoref{tab:breast} and \autoref{tab:retina} as we enter into datasets having considerably fewer available samples (Breast and Retina having $780$ and $1600$ respectively compared to Pneumonia's $5856$ samples).In both these datasets, the non-contrastive method (VICReg in most cases) emerges with the best performance under almost every federated setup examined. We also observe each algorithm having very high standard deviation in its performance across all $3$ datasets, implying high variance among the individual clients. This can be attributed to the non-i.i.d. distribution of samples in a low data setting leading to  clients having severely smaller and possibly more imbalanced datasets and thus, higher variations in performance.

\keypoint{Big datasets:}From tables \ref{tab:organA}, \ref{tab:organC}, \ref{tab:organS}, we once again see VICReg emerging as the best performer overall. SimCLR too shows considerably better performance on these datasets compared to the smaller ones. We attribute this improvement to the greater amount of data available in these $3$ datasets (The $3$ OrganMNIST datasets have nearly $10$ times more data compared to the smaller datasets like Breast or Retina), which generally aids all self-supervision methods. Another important observation from tables \ref{tab:organA}, \ref{tab:organC}, \ref{tab:organS} is the significantly lower standard deviation of $F1-scores$ of each algorithm relative to those in tables \ref{tab:pneumonia}, \ref{tab:breast}, \ref{tab:retina}, which indicate a lower variance of scores among the clients. With the Organ datasets having a considerably large amount of data, each client is left with a larger and potentially more balanced set of samples, even under non-i.i.d. distribution.

%-----------------%

\subsection{Across FL algorithms}
\begin{figure*}[h]
    \captionsetup[subfigure]{labelformat=empty}
    \centering
    \resizebox{\textwidth}{!}{
    \subfloat[]{\includegraphics{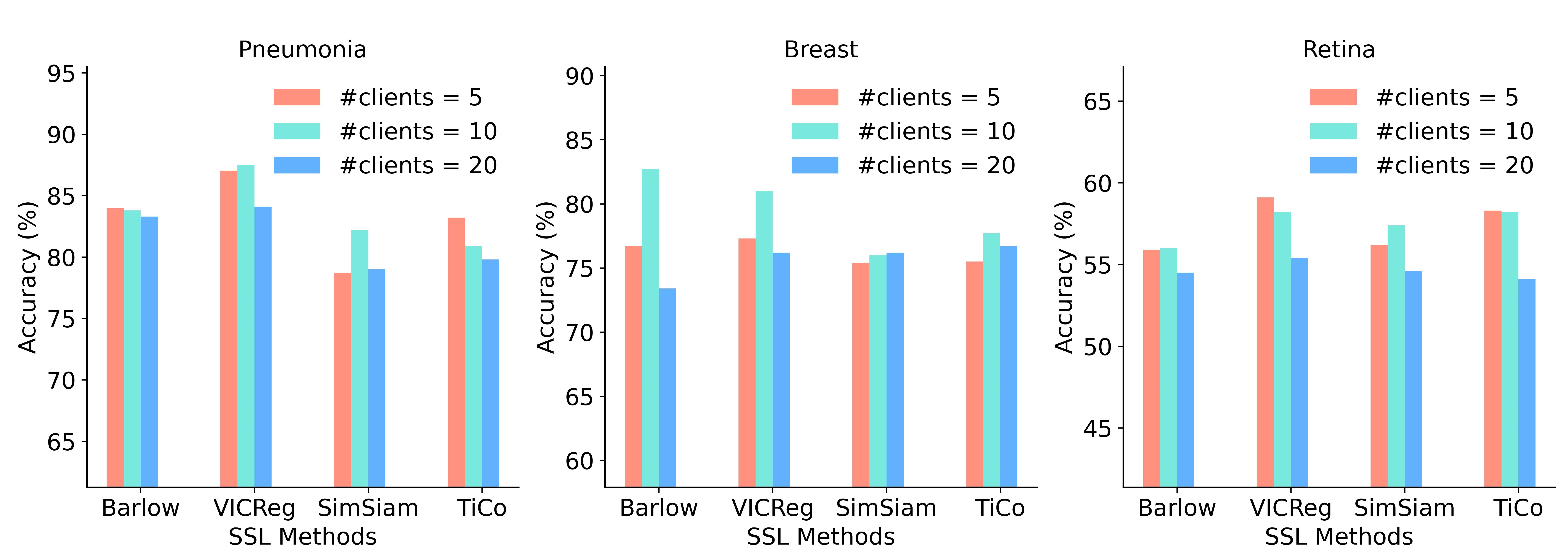}}\;\;
    \subfloat[]{\includegraphics{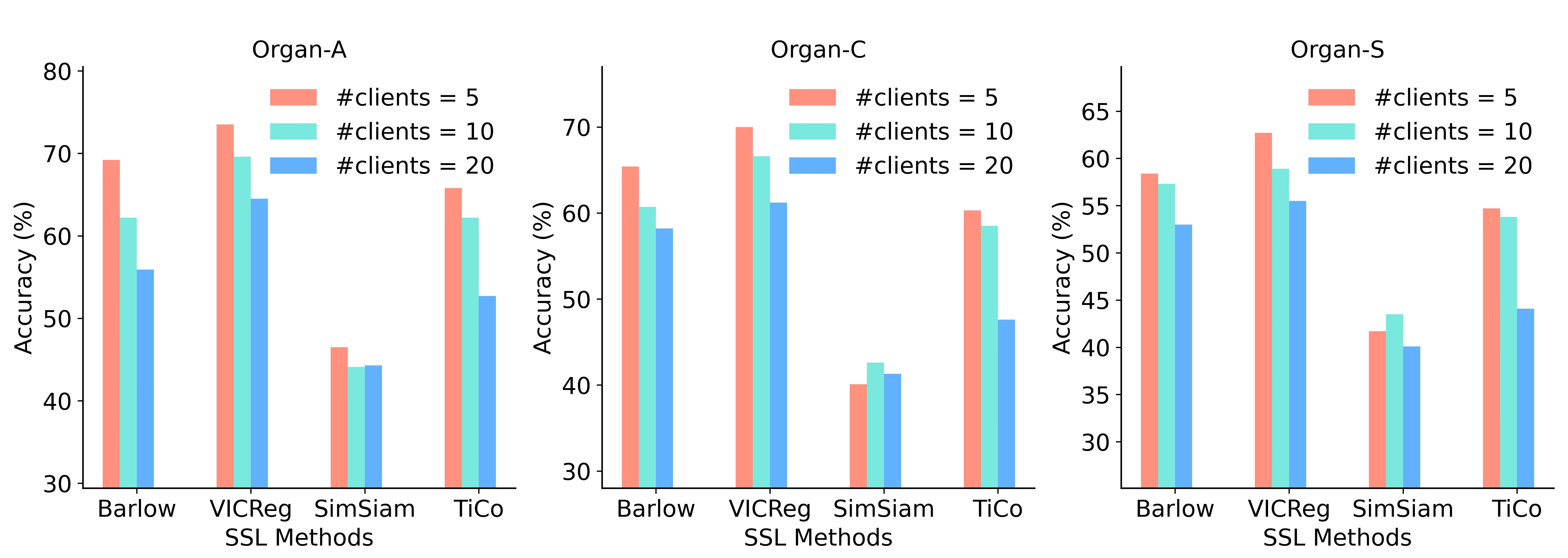}}\;\;
    }
    \caption{Performance across various number of clients under FedAVG simulation.}
    \label{fig:FedAVG_clients}
\end{figure*}

\begin{figure*}[h]
    \captionsetup[subfigure]{labelformat=empty}
    \centering
    \resizebox{\textwidth}{!}{
    \subfloat[]{\includegraphics{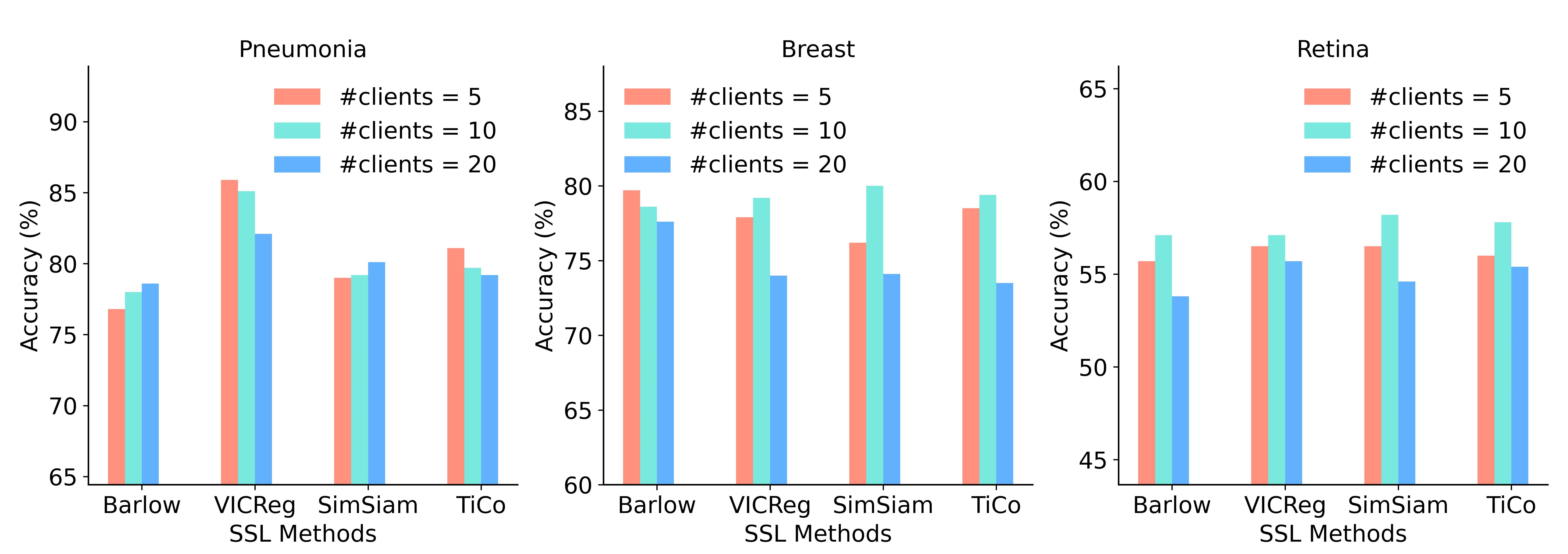}}\;\;
    \subfloat[]{\includegraphics{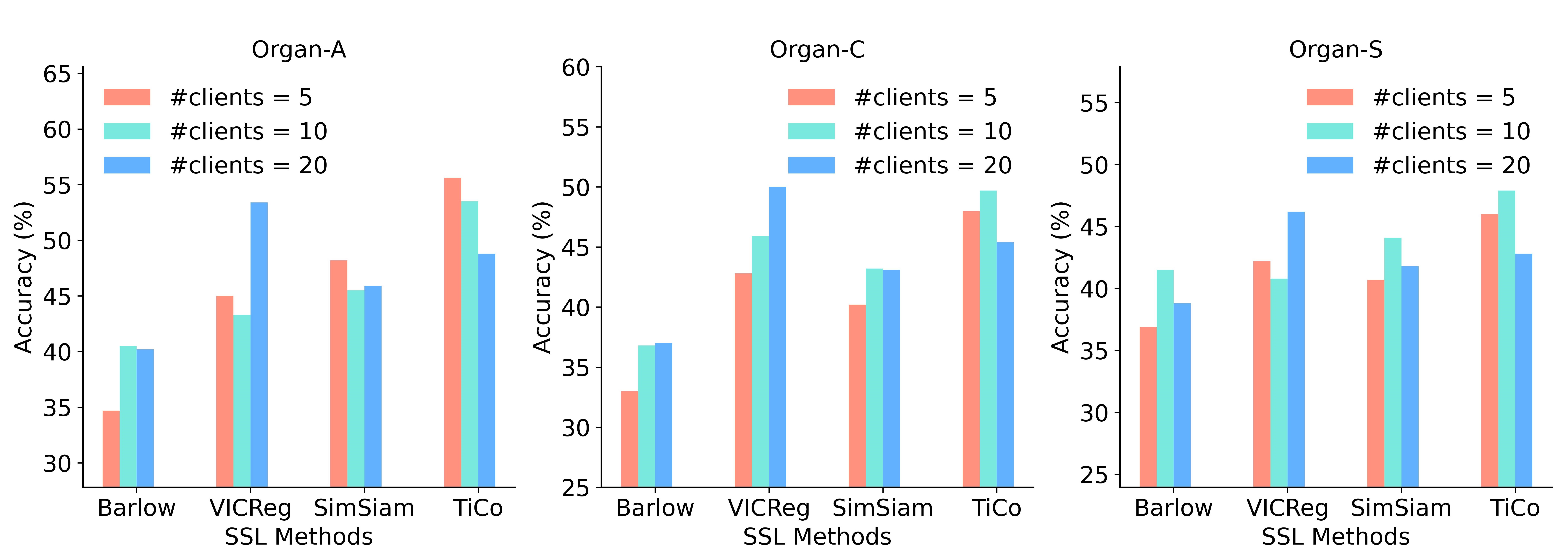}}\;\;
    }
    \caption{Performance across various number of clients under FedBN simulation.}
    \label{fig:fedbn_clients}
\end{figure*}

\begin{figure*}[!h]
    \captionsetup[subfigure]{labelformat=empty}
    \centering
    \resizebox{\textwidth}{!}{
    \subfloat[]{\includegraphics{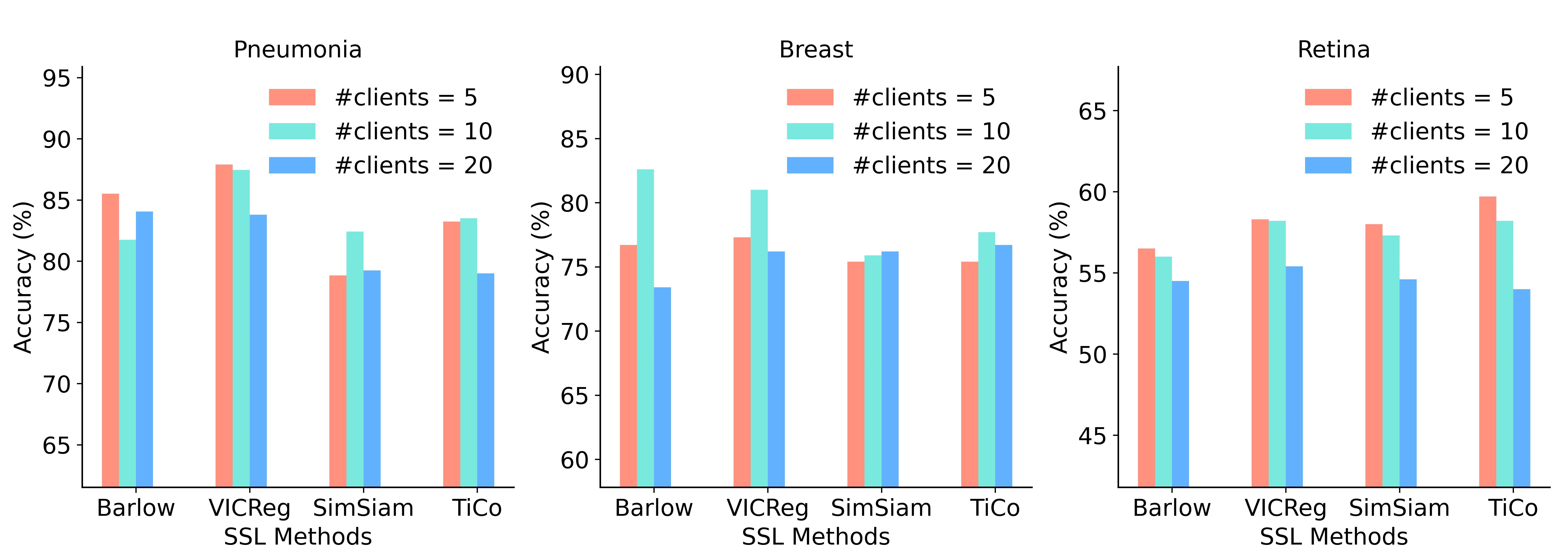}}\;\;
    \subfloat[]{\includegraphics{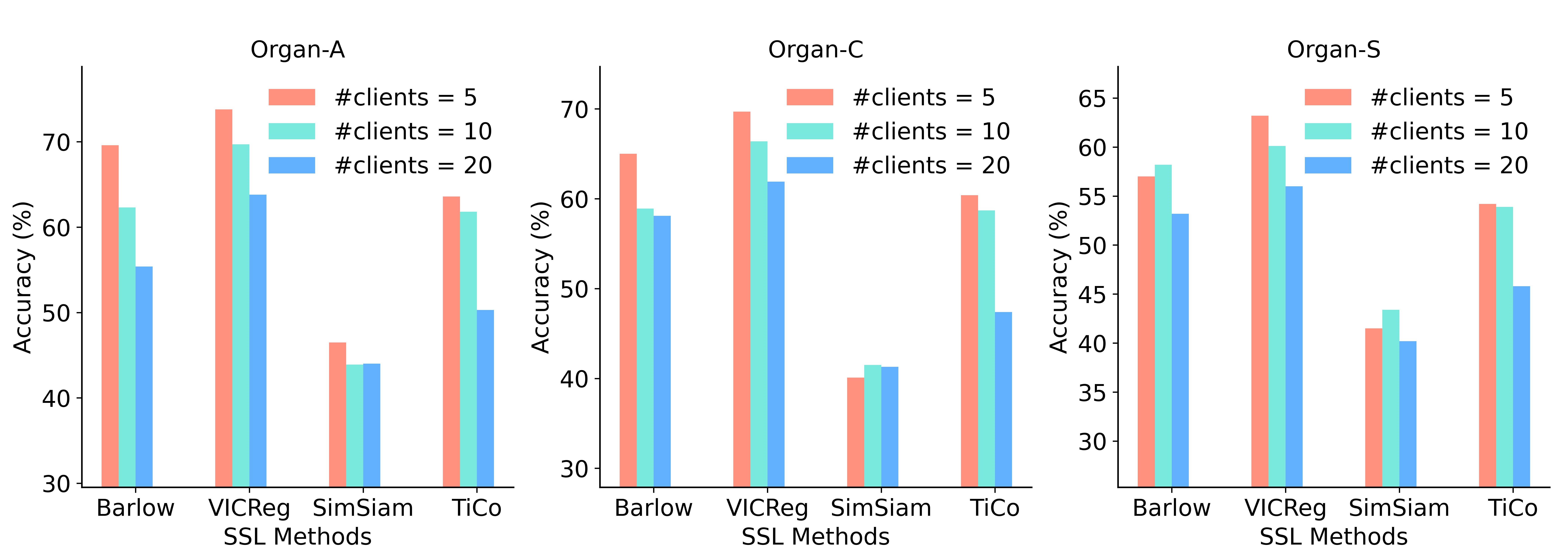}}\;\;
    }
    \caption{Performance across various number of clients under FedProx simulation.}
    \label{fig:fedprox_clients}
\end{figure*}

We studied the behaviour of non-contrastive methods subjected under the different federated learning simulations -- FedAVG \cite{mcmahan2017communication}, FedBN \cite{li2021fedbn} and FedProx \cite{li2020fedprox}. The findings have been graphically presented in \autoref{fig:FedAVG_clients}, \autoref{fig:fedbn_clients} and \autoref{fig:fedprox_clients} respectively.

%%% NCSSL performance with varying nos of clients %%%

\keypoint{FedAVG:}\autoref{fig:FedAVG_clients} shows the performances of each non-contrastive SSL variant across various numbers of clients under FedAVG \cite{mcmahan2017communication} scheme. We observe that there is a general trend of performance deterioration with increase in clients. This is expected, as more clients imply lesser data in each of them along with a higher percentage of imbalanced clients due to non-i.i.d. settings. For the OrganMNIST datasets, the behaviours of Barlow \cite{zbontar2021barlow}, VICReg \cite{bardes2021vicreg} and TiCo \cite{zhu2022tico} are very similar in this regard, with VICReg being the best performer among a majority of cases. On the other hand, among the ``small'' datasets (Pneumonia, Breast and Retina), a majority of the algorithms peak in performance when utilising $10$ clients, followed by a sharp dip when the number of clients is further increased further to $20$. This may be attributed to the severe decrease in the local client data (as the datasets are even smaller; $\approx 10$x smaller than the Organ datasets) which hurts the learning ability of self-supervision models in general \cite{chen2021exploring, chen2021intriguing}.

\keypoint{FedBN:}As evident from \autoref{fig:fedbn_clients}, FedBN simulation results are in contrast to the expected behaviour, (i.e., with increase in $\#clients$ the performance should drop). For VICReg, the general trend across the \textit{small} datasets is the aforementioned expected behaviour, however, for \textit{big} datasets such as Organ-C, we see an increase in the accuracy values with an increase in $\#clients$. It is also important to note that out of the 4 Non-CL algorithms in \autoref{fig:fedbn_clients}, TiCo \cite{zhu2022tico} shows the \textit{most} consistent general trend across all the datasets. Nonetheless, from the above observations we can conjecture that local Batch Normalization plays a pivotal role in the variability of the performances of SSL algorithms for the datasets taken into consideration.

\keypoint{FedProx:}Comparing \autoref{fig:FedAVG_clients} and \autoref{fig:fedprox_clients}, it can be inferred that FedProx shows very similar trends to FedAVG across a varying number of clients, which is reasonable as it only makes lightweight alterations to the latter in terms of introducing the proximal regularization term \cite{li2020fedprox}. For each SSL method, accuracy values obtained under FedAVG and FedProx are very close to each other, with minor performance gains observed for TiCo \cite{zhu2022tico} and VICReg \cite{bardes2021vicreg} under FedProx simulation scores.

%-----------------%

\subsection{Contrastive vs Non-Contrastive SSL}

\begin{figure*}[h]
    \captionsetup[subfigure]{labelformat=empty}
    \centering
    \resizebox{\textwidth}{!}{
    \subfloat[]{\includegraphics{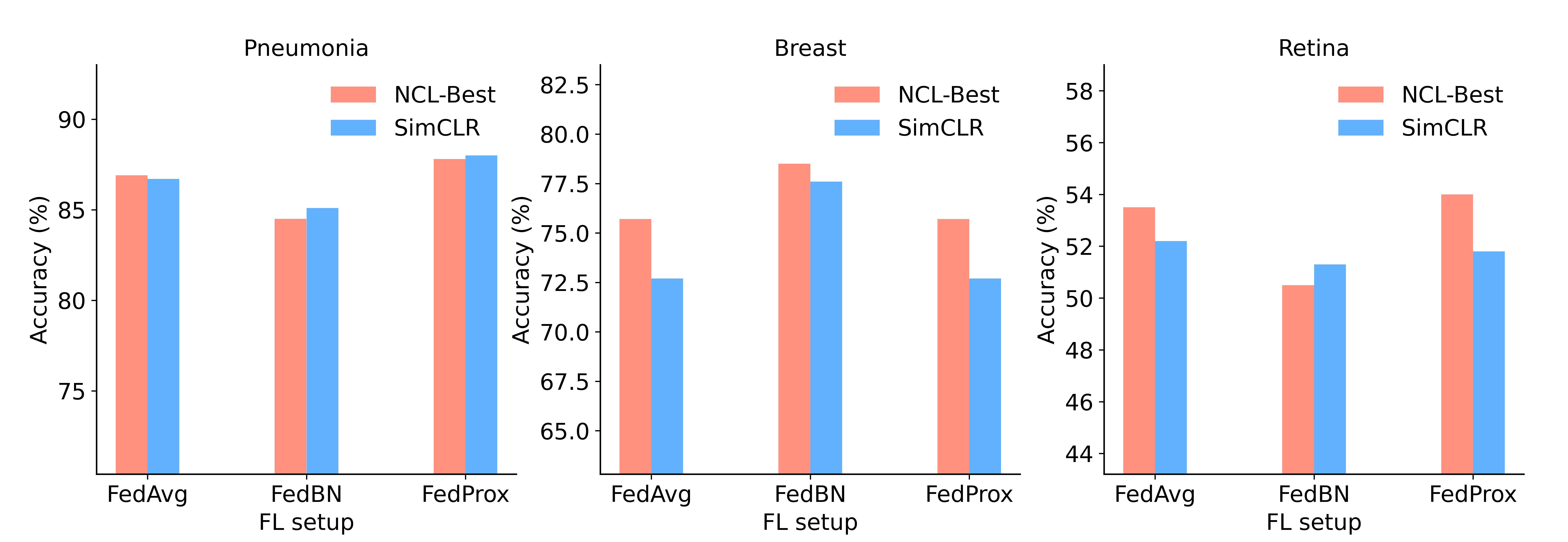}}\;\;
    \subfloat[]{\includegraphics{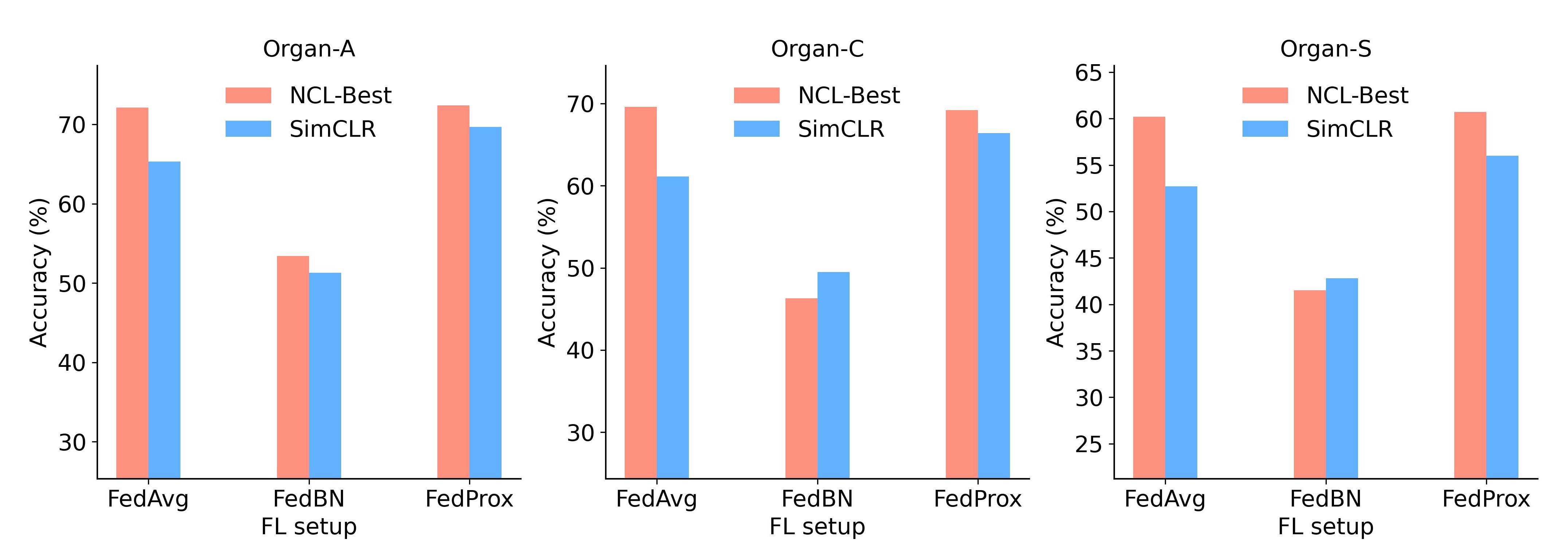}}\;\;
    }
    \caption{Contrastive vs non-contrastive for $\#clients=5$.}
    \label{fig:client5}
\end{figure*}

\begin{figure*}[h]
    \captionsetup[subfigure]{labelformat=empty}
    \centering
    \resizebox{\textwidth}{!}{
    \subfloat[]{\includegraphics{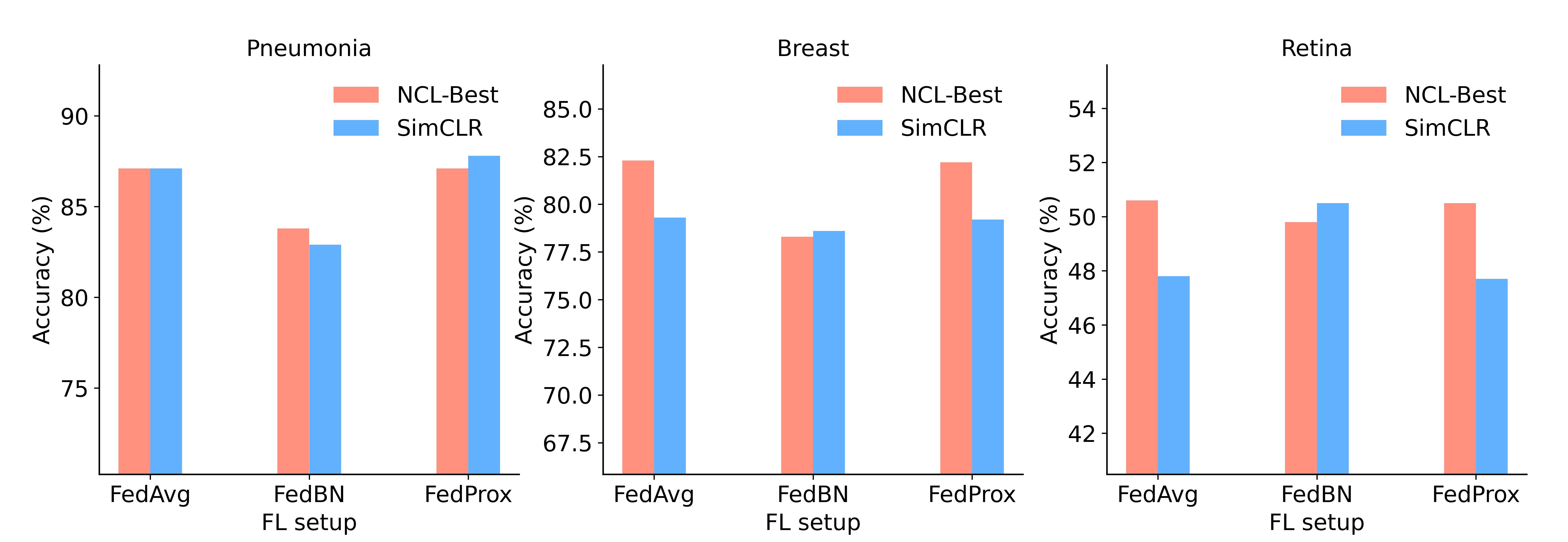}}\;\;
    \subfloat[]{\includegraphics{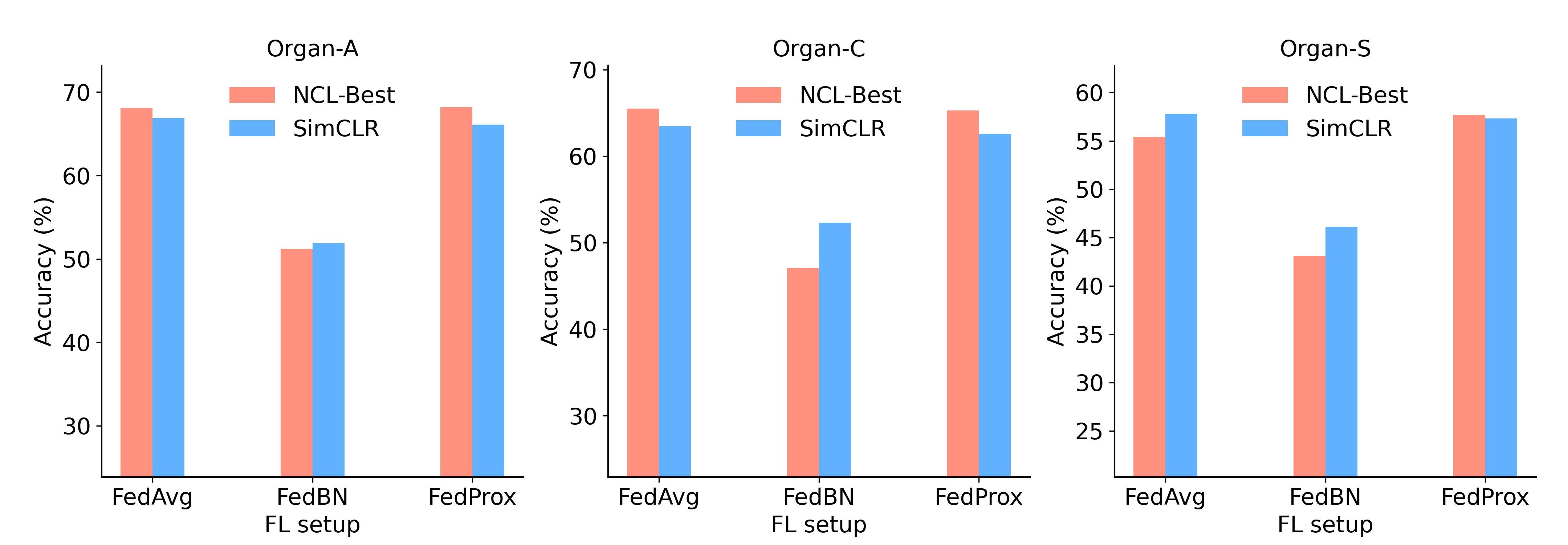}}\;\;
    }
    \caption{Contrastive vs non-contrastive for $\#clients=10$.}
    \label{fig:client10}
\end{figure*}

\begin{figure*}[!h]
    \captionsetup[subfigure]{labelformat=empty}
    \centering
    \resizebox{\textwidth}{!}{
    \subfloat[]{\includegraphics{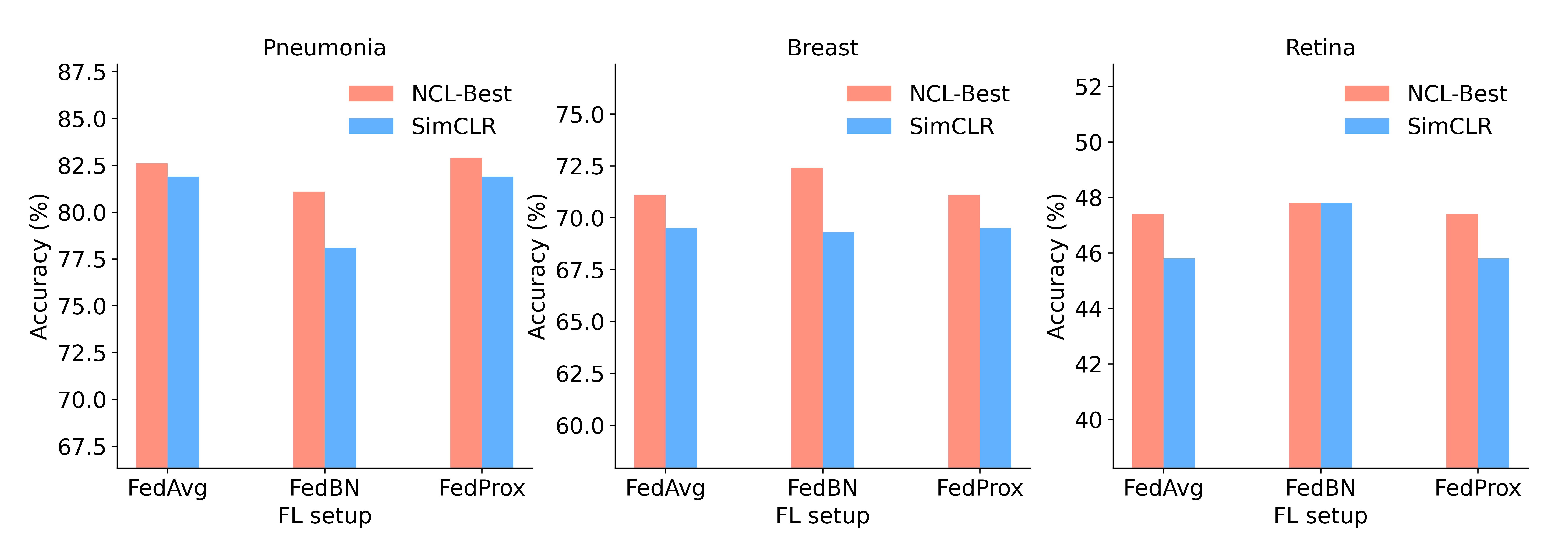}}\;\;
    \subfloat[]{\includegraphics{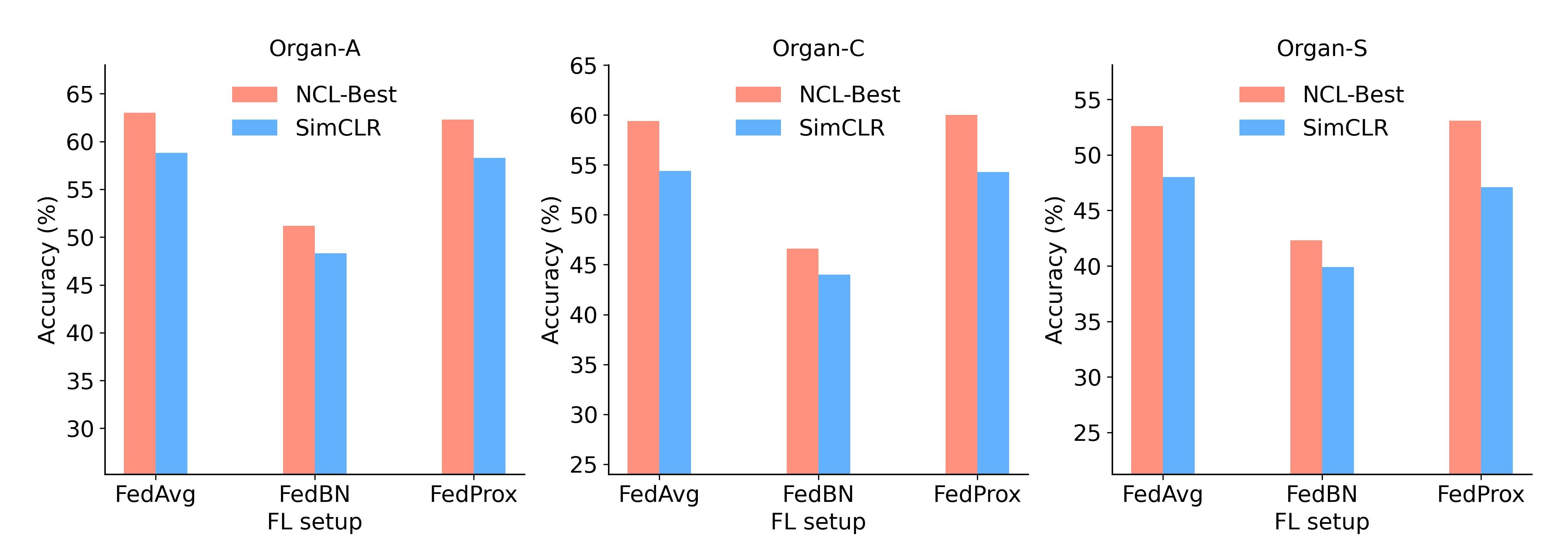}}\;\;
    }
    \caption{Contrastive vs non-contrastive for $\#clients=20$.}
    \label{fig:client20}
\end{figure*}

Since our study intends to investigate the behaviour of non-contrastive SSL variants under FL setups, we seek to analyse how they fare against the contrastive SimCLR \cite{chen2020simple} -- a strong self-supervision baseline for image classification tasks \cite{azizi2021big, shurrab2022self, jaiswal2020survey}. To do so, for each of the experimental setups we compare the accuracy obtained by the best-performing non-contrastive method (denoted by NCL-Best) with that of SimCLR. The plots depicting these comparisons have been provided in \autoref{fig:client5}, \autoref{fig:client10} and \autoref{fig:client20}, distributed according to the number of clients considered ($5, 10$ and $20$). The reason for such a comparison lies in the fact that it is not possible to find a \emph{single best} non-contrastive method for all setups, as well as to make our study more generalised rather than specific to a particular SSL algorithm. 

%%% NCL vs SimCLR for different fed algos %%%

\keypoint{Number of clients:}From the figures, it is evident that NCL-Best outperforms SimCLR on almost all setups, with several having significant margins of difference. The trend is best prominent in \autoref{fig:client20} ($\#clients=20$), where the batch size is reduced to $64$ (refer to \autoref{tab:hp} to cater to the low data regime caused by the increased number of clients. It is well-known \cite{chen2020simple, chen2021intriguing} that contrastive approaches work better with higher batch sizes (or greater amount of data \cite{he2020momentum}), due to the availability of more negative samples. Moreover, from the plots, it is clear that the performance of SimCLR decreases with an increase in clients (i.e. decrease in effective data sizes across clients). Whereas non-contrastive methods being negative-free do not have such a dependency and thus are more stable across the number of clients, as well as expectedly beating the former in the low-data regime. With lesser number of clients too (\autoref{fig:client5} and \autoref{fig:client10}), non-contrastive methods mostly prove to be more effective than SimCLR, but the trend is less prominent compared to those in \autoref{fig:client20}. One noteworthy observation here is, that under FedBN \cite{li2021fedbn} setup across a lower number of clients, the contrastive approach seems to be more favourable compared to the non-contrastive ones, which is \emph{contrary} to the overall behaviour observed. However, in the low data regime (\autoref{fig:client20}), it follows the general trend.

\keypoint{Dataset size:}We revisit figures \ref{fig:client5}, \ref{fig:client10} and \ref{fig:client20}, this time keeping in mind the dataset sizes (as previously described in Section \ref{sec:data}). We observe that among ``small'' datasets, the difference between NCL-Best and SimCLR is smaller compared to those in ``big'' datasets. The gaps, however, increase with the number of clients and make the trends discussed above more generalised. Furthermore, SimCLR shows greater stability on the OrganMNIST datasets compared to the smaller datasets (Pnuemonia, Breast and Retina) when $\#clients$ is varied.

From these observations, it is reasonable to infer that non-contrastive self-supervised methods \cite{zbontar2021barlow, bardes2021vicreg, zhu2022tico} are more effective than contrastive ones under federated simulations (especially under low data availability) and thus, are worthy of greater attention and future research.

%----------------------%

%--------------------%

\subsection{Further analysis}

\keypoint{SimSiam performs poorly on OrganMNIST:}Studying \autoref{fig:FedAVG_clients} and \autoref{fig:fedprox_clients}, it can be clearly seen that SimSiam performs much poorer than its counterparts on the 3 OrganMNIST datasets. As discussed in \cite{li2022understanding}, this can be attributed to SimSiam succumbing to dimensional collapse resulting from relatively smaller size of encoder (ResNet-18) and comparably higher data complexity ($\approx 10$x) in the 3 larger OrganMNIST datasets. This is further borne out by the significant performance gain we see in \autoref{fig:fedbn_clients} caused by a reduction in the local bias via exclusion of batch normalization layers from global model updates in FedBN \cite{li2021fedbn}.
% \keypoint{Performance drop of SimSiam for larger datasets:} Another observation that we noted from our experimental results is that the SimSiam Algorithm tends to perform poorly for large sized datasets with respect to other contrastive and non contrastive algorithms that we have used. This observation is in accordance with the observations obtained from the \cite{li2022understanding}. However we also noted that performance of the SimSiam algorithm matches with other algorithms for small datasets like Pneumonia, Breast and Retina. 

\keypoint{FedBN aids SimCLR:}In \autoref{fig:client5} and \autoref{fig:client10}, we observe that SimCLR seems to perform comparatively better (with respect to the best non-contrastive method) under FedBN compared to the other FL algorithms. While it still doesn't beat the best-performing non-contrastive approach in all instances, the gap in performance is significantly smaller when FedBN is used. However, we also note that in \autoref{fig:client20}, FedBN behaves in a similar fashion as FedAVG or FedProx, thus suggesting that the underlying cause maybe due to the batch sizes considered. We point this out as an intriguing observation which may be investigated in future studies.

% \keypoint{Contrastive vs Non-contrastive SSL:}

%--------------------------Graph Section-------------------------------------
% ------------------------------------------------------------------------------

% ------------------------------------------------------------------------------

\section{Conclusion}
Despite the great progress in medical image analysis, the concerns regarding privacy and annotation bottleneck are yet to be fully resolved. In this study, we take a step forward and tackle these challenges head-on by combining federated learning with self-supervision for medical image classification. We experimented with SoTA SSL algorithms exhaustively across various FL setups, simulated by varying the number of clients, to conduct realistic non-i.i.d simulations on the MedMNIST data suite. Our results suggest the high applicability of non-contrastive SSL methods to such tasks, which are found to outperform the contrastive baseline by fair margins. With a systematic analysis of our findings, we show the trends of different algorithms and simulations across different dataset sizes, number of clients, etc. Our holistic evaluation and benchmarking is the first of its kind, which we feel should be of great assistance to other researchers working in related domains. The inferences drawn from our findings, as well as some of the ``anomalies'' (e.g. surprisingly poor behaviour of SimSiam, contrasting trends of FedBN), should be of great interest for fellow researchers for further exploration. In future, we plan to expand the horizon of our investigations by digging deeper into the nuances of each FL setting to better interpret the observed trends.

%%%%%%%%% REFERENCES
{\small
\bibliographystyle{ieee_fullname}
\bibliography{biblio}
}

\end{document}